\documentclass{article}

\usepackage{microtype}
\usepackage{graphicx}
\usepackage{subcaption}
\usepackage{booktabs} % for professional tables

\usepackage{hyperref}

\usepackage[preprint]{icml2026}

\usepackage{amsmath}
\usepackage{amssymb}
\usepackage{mathtools}
\usepackage{amsthm}

\usepackage{microtype}
\usepackage{graphicx}
\usepackage{subcaption}
\usepackage{booktabs} % for professional tables
\usepackage{multirow}     % \multirow
\usepackage[table]{xcolor}% \cellcolor and table color support
\usepackage{makecell}          % 支持单元格换行
\usepackage{threeparttable}   % 支持表格注释
\usepackage{adjustbox}         % 自动缩放表格
\usepackage{wrapfig}           % 图文环绕
\usepackage{arydshln}
\usepackage{colortbl,xcolor}   % 表格着色
\definecolor{color3}{rgb}{0.95,0.95,0.95} % 浅灰背景色
\definecolor{ssr}{HTML}{43901a}  % 深绿色，保留不变
\definecolor{atq}{HTML}{bc8900}  % 暗金色，更新为新的颜色
\usepackage{amsmath, amssymb} % 用于提供小三角符号
\usepackage[most]{tcolorbox}

\usepackage{amsmath, amssymb}

\usepackage{amsthm}

\usepackage{xcolor}

\newcommand{\mycomment}[1]{\hfill\small\texttt{$\triangleright$ #1}}

\usepackage[capitalize,noabbrev]{cleveref}

\theoremstyle{plain}

\theoremstyle{definition}

\theoremstyle{remark}

\usepackage[textsize=tiny]{todonotes}

\icmltitlerunning{}

\begin{document}

\twocolumn[
  % \icmltitle{D$^2$Quant: Towards Accurate Post-Training Weight Quantization for LLMs}
  \icmltitle{D$^2$Quant: Accurate Low-bit Post-Training Weight Quantization for LLMs}


  \icmlsetsymbol{equal}{*}

  \begin{icmlauthorlist}
    \icmlauthor{Xianglong Yan}{equal,yyy}
    \icmlauthor{Chengzhu Bao}{equal,yyy}
    \icmlauthor{Zhiteng Li}{yyy}
    \icmlauthor{Tianao Zhang}{yyy}
    \icmlauthor{Shaoqiu Zhang}{yyy} \\
    \icmlauthor{Ruobing Xie}{comp}
    \icmlauthor{Xingwu Sun}{comp}
    \icmlauthor{Yulun Zhang\textsuperscript{\textdagger}}{yyy}
  \end{icmlauthorlist}

  \icmlaffiliation{yyy}{Shanghai Jiao Tong University}
  \icmlaffiliation{comp}{Tencent Hunyuan}

  \icmlcorrespondingauthor{Yulun Zhang}{yulun100@gmail.com}

  \icmlkeywords{Machine Learning, ICML}

  \vskip 0.3in
]



\printAffiliationsAndNotice{\icmlEqualContribution}

\begin{abstract}
Large language models (LLMs) deliver strong performance, but their high compute and memory costs make deployment difficult in resource-constrained scenarios. Weight-only post-training quantization (PTQ) is appealing, as it reduces memory usage and enables practical speedup without low-bit operators or specialized hardware. However, accuracy often degrades significantly in weight-only PTQ at sub-4-bit precision, and our analysis identifies two main causes: (1) down-projection matrices are a well-known quantization bottleneck, but maintaining their fidelity often requires extra bit-width; (2) weight quantization induces activation deviations, but effective correction strategies remain underexplored. To address these issues, we propose D$^2$Quant, a novel weight-only PTQ framework that improves quantization from both the weight and activation perspectives. On the weight side, we design a Dual-Scale Quantizer (DSQ) tailored to down-projection matrices, with an absorbable scaling factor that significantly improves accuracy without increasing the bit budget. On the activation side, we propose Deviation-Aware Correction (DAC), which incorporates a mean-shift correction within LayerNorm to mitigate quantization-induced activation distribution shifts. Extensive experiments across multiple LLM families and evaluation metrics show that D$^2$Quant delivers superior performance for weight-only PTQ at sub-4-bit precision. The code and models will be available at \url{https://github.com/XIANGLONGYAN/D2Quant}. 
\end{abstract}

%% narrow the gap between equations and sentences
\setlength{\abovedisplayskip}{2pt}
\setlength{\belowdisplayskip}{2pt}

\vspace{-9mm}
\section{Introduction}
\vspace{-2mm}
Large language models (LLMs)~\citep{grattafiori_llama_2024,qwen3technicalreport,zhang_opt_2022} have achieved remarkable success in natural language processing (NLP), demonstrating strong capabilities in both understanding and generation. Yet, this progress is largely driven by scaling up model size. Consequently, modern LLM families such as Qwen~\citep{qwen3technicalreport} and LLaMA~\citep{grattafiori_llama_2024} continue to grow to further improve performance. However, the high memory and compute costs of LLM inference make deployment challenging in resource-constrained settings, limiting their real-world use on edge and mobile devices.

\begin{figure}[t!]
    \centering
    \includegraphics[width=\linewidth]{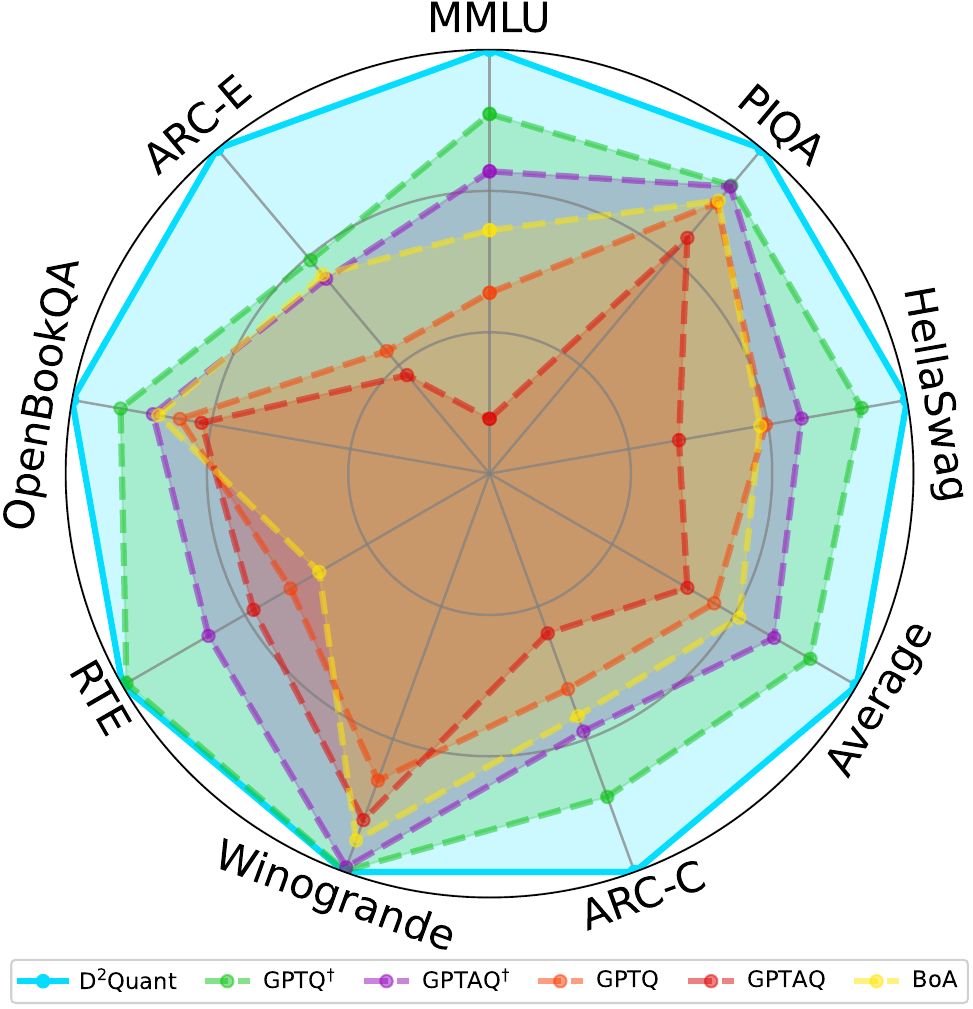}
    \vspace{-4.5mm}
    \caption{Performance comparison of weight-only PTQ methods on Qwen-3-8B with 2-bit quantization. D$^2$Quant consistently outperforms all other methods across all evaluation metrics.}
    \vspace{-9mm}
    \label{fig:fig1}
\end{figure}

\begin{figure*}[t]
\includegraphics[width=1\textwidth]{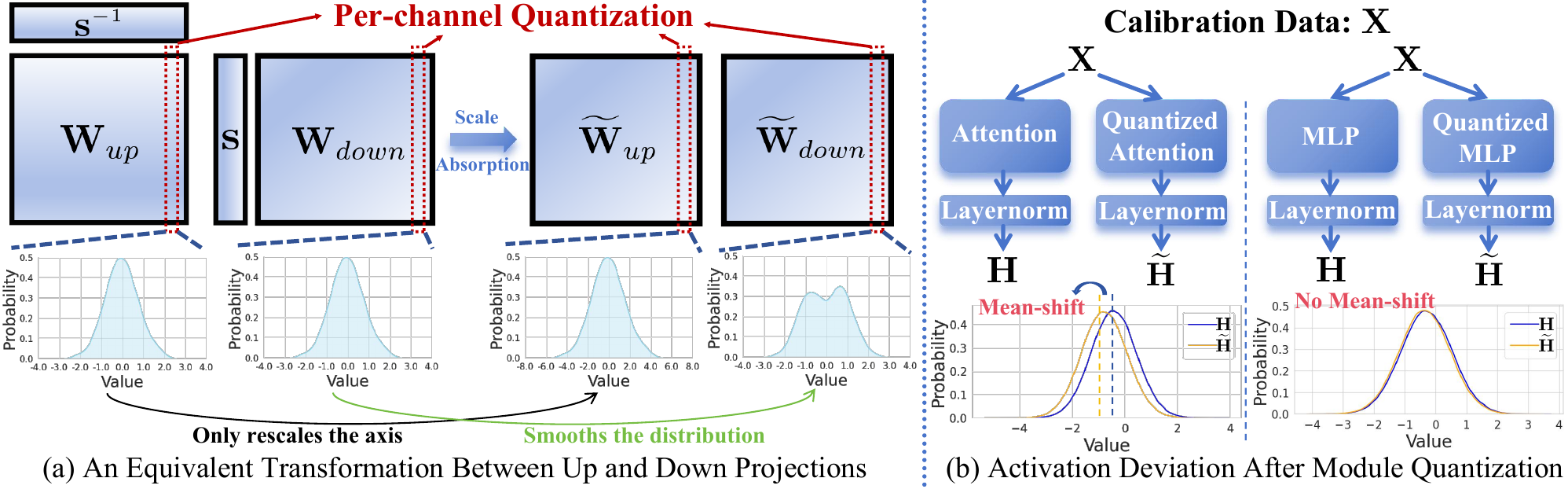}
\vspace{-6mm}
\caption{
\textbf{(a)} Equivalent transformation between up and down projections in per-channel quantization: smoothing can be applied to down projection, while up projection only introduces channel-wise scaling. \textbf{(b)} Activation deviation at the subsequent LayerNorm after quantizing attention and MLP: quantizing attention causes a notable mean shift, whereas MLP quantization introduces no significant shift.
}
\label{fig:fig2}
\vspace{-6.5mm}
\end{figure*}

Quantization is one of the most effective ways to compress LLMs and enable deployment. 
In conventional neural networks~\citep{krishnamoorthi2018quantizing,esser_lsq_2019}, quantization is often applied to both weights and activations, and low-bit matrix multiplication is used to reduce compute and speed up inference. However, these gains typically depend on specialized operators and hardware support. In contrast, LLM inference is largely memory-bound, which makes weight-only quantization especially attractive. By compressing weights alone, it not only reduces the overall memory footprint, but also enables practical inference speedups by alleviating memory bandwidth bottlenecks, without requiring low-bit operators or specialized hardware.

Weight-only quantization for LLMs typically follows two paradigms: quantization-aware training (QAT) and post-training quantization (PTQ). In practice, PTQ is more widely adopted as it avoids retraining and thus requires significantly fewer computational and data resources. Representative PTQ methods, such as GPTQ~\citep{frantar_gptq_2023} and AWQ~\citep{lin_awq_2024}, perform well at 4-bit. However, reducing the bit-width below 4 often leads to a pronounced accuracy drop (see Fig.~\ref{fig:fig1}). To improve sub-4-bit PTQ, we make two key observations and analyses:
% \vspace{-6mm}
\begin{itemize}
\vspace{-3.5mm}
\item It is widely recognized that down-projection matrices are highly sensitive to quantization. As shown in Fig.~\ref{fig:fig2}(a), we find that inserting an equivalent scaling transformation between the up- and down-projections is beneficial in a weight-only setting. It makes the down-projection easier to quantize while leaving the up-projection’s quantization difficulty unchanged. This offers a principled way to improve down-projection accuracy without increasing the bit budget.
\vspace{-2mm}
\item Although activations are not quantized in weight-only PTQ, they can still drift due to unavoidable weight quantization errors, which can severely affect model outputs. We measure activation deviations after quantizing the attention and MLP blocks.  As shown in Fig.~\ref{fig:fig2}(b), we observe a clear mean shift at the post-attention LayerNorm after quantizing the attention module, while this behavior is much less evident after quantizing the MLP module. This pronounced discrepancy provides a useful cue for correcting activation drift in weight-only quantization.
\end{itemize}
\vspace{-4mm}

Based on these observations, we propose D$^2$Quant, a weight-only PTQ framework that improves sub-4-bit quantization from both weight and activation perspectives. \textbf{From the weight perspective}, we design a Dual-Scale Quantizer (DSQ), which reformulates the smoothing between up- and down-projection as a dual-scale quantization problem on the down-projection. By incorporating the additional scaling factor into the down-projection quantization process, we develop an efficient optimization scheme that enables the two scaling factors at different granularities to rapidly converge to their optimal values. Notably, the additional scaling factor can be fully folded into the preceding up-projection after quantization, improving down-projection fidelity with essentially no extra bit budget or inference overhead. \textbf{From the activation perspective}, we first introduce a \emph{signal-to-noise ratio} to quantify activation drift, formalizing our earlier observation that attention quantization induces a pronounced mean shift at the post-attention LayerNorm, whereas the pre-LayerNorm exhibits less structured deviation after MLP quantization. Motivated by this, we propose Deviation-Aware Correction (DAC), which injects a lightweight deviation correction term into the post-attention LayerNorm to compensate for the quantization-induced mean shift. We further provide a theoretical analysis showing that the expected error reduction achieved by DAC is directly related to the \emph{signal-to-noise ratio} defined above.

Extensive experiments across multiple LLM families and evaluation metrics show that D$^2$Quant delivers superior performance for weight-only PTQ at sub-4-bit precision. As shown in Fig.~\ref{fig:fig1}, on Qwen3-8B under 2-bit quantization, D$^2$Quant achieves an average accuracy of 57.22 over seven zero-shot tasks (vs. 54.05 for the state of the art (SOTA)).

\begin{figure*}[t]
\includegraphics[width=1\textwidth]{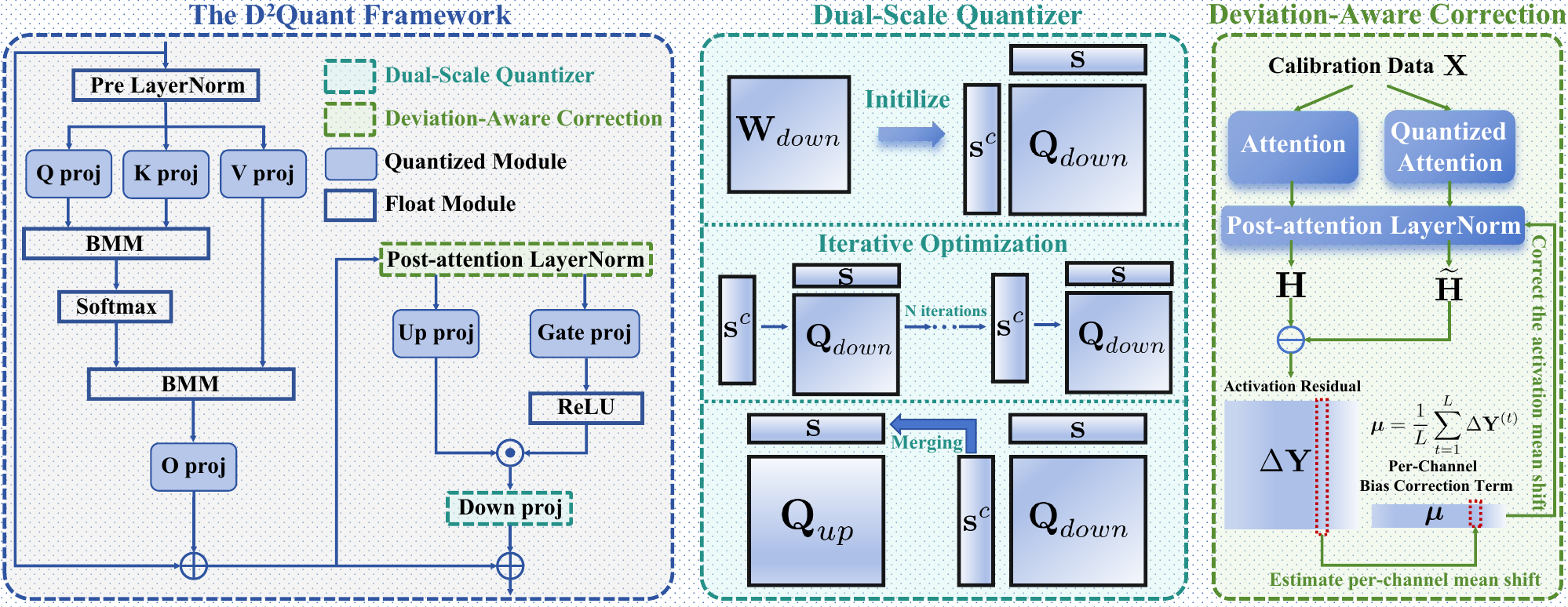}
\vspace{-5mm}
\caption{
Overview of \textbf{D$^2$Quant}. The left panel shows the D$^2$Quant framework, which improves weight-only PTQ at both weight and activation levels. The middle shows the \emph{Dual-Scale Quantizer}, which introduces an additional scale to refine the down-projection. The right depicts the \emph{Deviation-Aware Correction}, which mitigates mean shift at post-attention LayerNorm via bias alignment.
}\label{fig:fig3}
\vspace{-4mm}
\end{figure*}

Our main contributions are summarized as follows: 
\vspace{-2mm}
\begin{itemize}
\vspace{-1mm}
\item At the weight level, we design a \textbf{Dual-Scale Quantizer (DSQ)} tailored to down-projection weight matrices, optimizing an absorbable auxiliary scaling factor to improve accuracy without increasing the bit budget.
\vspace{-1mm}
\item At the activation level, we propose \textbf{Deviation-Aware Correction (DAC)}, which performs mean-shift correction in the post-attention LayerNorm to mitigate activation drift, yielding improved model performance.
\vspace{-5mm}
\item By combining DSQ and DAC, we develop \textbf{D$^2$Quant}, which achieves superior sub-4-bit performance for weight-only PTQ, effectively mitigating severe performance degradation in this regime.
\end{itemize}

\vspace{-1mm}
\section{Related Works}
\vspace{-1mm}
\subsection{Large Language Model Quantization}
\vspace{-1mm}
Current LLM quantization techniques can be broadly categorized into quantization-aware training (QAT) and post-training quantization (PTQ). \textbf{QAT}~\citep{shao_omniquant_2024,du_bitdistiller_2024,ashkboos_halo_2025,liu2025paretoq} integrates quantization into training, enabling the model to adapt to low-precision representations. To alleviate the data barrier in QAT, LLM-QAT~\citep{liu_llm-qat_2023} introduces data-free distillation. EfficientQAT~\citep{chen_efficientqat_2025} improves training efficiency via a two-stage procedure. Moreover, several works push low-precision training to sub-2-bit regimes~\citep{Xu2024OneBitTE,wang_bitnet_2023,BinaryMoS}. However, QAT requires training and thus incurs substantial data and computational overhead, whereas \textbf{PTQ}~\citep{yao_zeroquant_2022,wei_qdrop_2023,lee_owq_2024,wu2025quantcache} is training-free and considerably more resource-efficient. Recent studies further demonstrate that PTQ can achieve strong performance~\citep{dettmers2022gpt3,zhao2024atom,lin2025qserve}. SmoothQuant~\citep{xiao_smoothquant_2024} introduces a smoothing operation that transfers the quantization difficulty from activations to weights, enabling near-lossless W8A8 quantization. A series of rotation-based methods~\citep{ashkboos_quarot_2024,hu2025ostquant,lin2024duquant,liu2024spinquant,sun2024flatquant} further smooth outliers in activations, advancing PTQ toward lower-bit regimes. Notably, most of the above PTQ approaches quantize both weights and activations for low-bit inference. We discuss weight-only PTQ, a key PTQ branch, in the next paragraph.
\vspace{-1mm}
\subsection{Weight-Only Post-training Quantization}
\vspace{-1mm}
Since LLM inference is typically memory-bound, compressing weights alone not only reduces memory footprint but also accelerates inference by reducing memory traffic. It requires no specialized low-bit operators or hardware support, making weight-only PTQ~\citep{dettmers_spqr_2023,kim_squeezellm_2024,chee_quip_2024,zhang2025quant,li_norm_2023} widely used in practical deployments. AWQ~\citep{lin_awq_2024} selectively protects salient weight channels based on activation statistics. GPTQ~\citep{frantar_gptq_2023} performs layer-wise quantization by minimizing output perturbation with an approximate Hessian. Building on GPTQ, GPTAQ~\citep{li_gptaq_2025} and QEP~\citep{arai_quantization_2026} account for input errors, while BOA~\citep{kim_boa_nodate} improves Hessian estimation, further boosting accuracy. Slim-LLM~\citep{huang_slim-llm_2025} adopts mixed precision with greedy bit allocation, assigning more bits to more important weights. Several methods~\citep{shang_pb-llm_2023,li2024arb,yan2025pt,yan2025progressive,huang2024billm} further push weight-only PTQ to sub-2-bit regimes. In addition, QuIP\#~\citep{tseng_quip_2024}, QTIP~\citep{tseng_qtip_2025}, GPTVQ~\citep{baalen_gptvq_2025}, and VPTQ~\citep{liu_vptq_2024} adopt vector quantization (VQ), grouping weights into vectors and quantizing them jointly to better capture inter-weight correlations at extremely low bit widths. However, performance often degrades sharply once the bit-width drops below 4, limiting the practicality of weight-only PTQ in this regime. Our work focuses on weight-only PTQ and aims to improve performance under sub-4-bit quantization.

\vspace{-0.5mm}
\section{Method}
% \vspace{-0.2mm}
In this section, we introduce D$^2$Quant, as illustrated in Fig.~\ref{fig:fig3}. First, Sec.~\ref{sec:preliminary} reviews low-bit quantization preliminaries and notation. Next, Sec.~\ref{sec:DSQ} presents the Dual-Scale Quantizer (DSQ) for quantizing down-projection matrices, followed by Sec.~\ref{sec:DAC}, which introduces Deviation-Aware Correction (DAC) to address activation drift. Finally, Sec.~\ref{sec:pipeline} combines DSQ and DAC into the D$^2$Quant pipeline.

% \newpage
\subsection{Preliminary}
\vspace{-1mm}
\label{sec:preliminary}
\noindent \textbf{Low-Bit Quantization.}
To facilitate efficient storage and high-speed inference, low-bit uniform quantization is utilized to map floating-point tensors into discrete low-precision representations. Formally, given a weight tensor $\mathbf{W} \in \mathbb{R}^{C_{\text{out}} \times C_{\text{in}}}$, we apply $b$-bit per-channel quantization to derive an integer tensor $\mathbf{W}_q \in \mathcal{Q}^{C_{\text{out}} \times C_{\text{in}}}$, where $\mathcal{Q} = \{0, 1, \dots, 2^b - 1\}$. The quantization is formulated as:
\begin{equation}
\mathbf{W}_q = \operatorname{clip} \left( \left\lfloor \frac{\mathbf{W}}{\mathbf{s}} \right\rceil + \mathbf{z}, \, 0, \, 2^b - 1 \right),
\label{eq:uniform_quant}
\end{equation}
where $\lfloor \cdot \rceil$ denotes the rounding-to-nearest-integer operator, and $\operatorname{clip}(\cdot)$ restricts the values to the range of $\mathcal{Q}$. The parameters $\mathbf{s} \in \mathbb{R}^{C_{\text{out}} \times 1}$ and $\mathbf{z} \in \mathcal{Q}^{C_{\text{out}} \times 1}$ represent the per-channel scale factors and zero-points, respectively, which are broadcast along the input-channel dimension. Specifically, the scale factor $\mathbf{s}$ and zero-point $\mathbf{z}$ are computed from the channel-wise dynamic range of $\mathbf{W}$ as:
\begin{equation} \mathbf{s} = \frac{\max(\mathbf{W}) - \min(\mathbf{W})}{2^b - 1}, \quad \mathbf{z} = -\left\lfloor \frac{\min(\mathbf{W})}{\mathbf{s}} \right\rceil.
\end{equation} 
Given the quantized integer tensor $\mathbf{W}_q$ along with the corresponding scale factors $\mathbf{s}$ and zero-points $\mathbf{z}$, we recover its floating-point approximation via dequantization:
\begin{equation}
\widehat{\mathbf{W}} = \mathbf{s} \odot (\mathbf{W}_q - \mathbf{z}),
\label{eq:uniform_dequant}
\end{equation}
where $\odot$ denotes element-wise multiplication and $\mathbf{s}, \mathbf{z}$ are broadcast to match the shape of $\mathbf{W}_q$.
% \vspace{-1mm}
% \noindent \textbf{LayerNorm.} 
% Layer Normalization (LayerNorm)~\citep{ba2016layer} is a core component in Transformer architectures. Given an input vector $\mathbf{X} \in \mathbb{R}^{d}$, LayerNorm is defined as:
% \begin{equation}
% \operatorname{LN}(\mathbf{X}) = \gamma \odot \frac{\mathbf{X}-\mu}{\sqrt{\frac{1}{d}\sum_{i=1}^{d}(X_i-\mu)^2}} + \beta,
% \end{equation}
% where $\mu=\frac{1}{d}\sum_{i=1}^{d}X_i$ is the feature-wise mean, and $\gamma,\beta$ are learnable scale and bias parameters. RMSNorm~\citep{zhang_rmsnorm_2019} simplifies LayerNorm by removing mean subtraction and normalizing only by the root mean square:
% \begin{equation}
% \operatorname{RMSNorm}(\mathbf{X}) = \gamma \odot \frac{\mathbf{X}}{\sqrt{\frac{1}{d}\sum_{i=1}^{d}X_i^2}},
% \end{equation}
% where $\gamma$ is a learnable scaling parameter. Since most modern LLMs (e.g., LLaMA-3~\citep{grattafiori_llama_2024} and Qwen-3~\citep{qwen3technicalreport}) adopt RMSNorm by default, we use "LayerNorm" to refer to RMSNorm throughout the rest of this paper unless otherwise specified.
\vspace{-3mm}
\subsection{Dual-Scale Quantizer} 
\vspace{-1mm}
\label{sec:DSQ}
\noindent \textbf{Equivalent Up--Down Scaling Transformation.}
The MLP module typically transforms the hidden states $\mathbf{X} \in \mathbb{R}^{L \times C_{\text{in}}}$ through a gated mechanism. Formally, given the input $\mathbf{X}$, the forward pass is defined as:
\begin{equation}
    \mathbf{Y} = \left( \sigma(\mathbf{X} \mathbf{W}_{\text{gate}}^\top) \odot (\mathbf{X} \mathbf{W}_{\text{up}}^\top) \right) \mathbf{W}_{\text{down}}^\top,
\end{equation}
where $\mathbf{W}_{\text{gate}}, \mathbf{W}_{\text{up}} \in \mathbb{R}^{H \times C_{\text{in}}}$ and $\mathbf{W}_{\text{down}} \in \mathbb{R}^{C_{\text{in}} \times H}$ denote the gate, up-projection, and down-projection matrices, respectively. A mathematically equivalent reparameterization can be introduced between the up- and down-projections via a per-channel scaling vector $\boldsymbol{\eta} \in \mathbb{R}^{1 \times H}$:
\begin{equation}
    \widetilde{\mathbf{W}}_{\text{up}}^\top = \mathbf{W}_{\text{up}}^\top \operatorname{diag}(\boldsymbol{\eta}), \,
    \widetilde{\mathbf{W}}_{\text{down}}^\top = \operatorname{diag}(\boldsymbol{\eta})^{-1} \mathbf{W}_{\text{down}}^\top,
\end{equation}
which results in the exact same computation:
\begin{equation}
    \mathbf{Y} = \left( \sigma(\mathbf{X} \mathbf{W}_{\text{gate}}^\top) \odot (\mathbf{X} \widetilde{\mathbf{W}}_{\text{up}}^\top) \right) \widetilde{\mathbf{W}}_{\text{down}}^\top.
\end{equation}
This transformation applies a shared scaling to the up- and down-projection matrices in opposite directions. Since both $\mathbf{W}_{\text{up}}$ and $\mathbf{W}_{\text{down}}$ are quantized with per-channel granularity, this transformation can have asymmetric effects: it can leave the quantization of $\mathbf{W}_{\text{up}}$ unaffected, as the scaling can be uniformly absorbed within each channel; meanwhile, it can smooth the distribution of $\mathbf{W}_{\text{down}}$, potentially reducing its dynamic range and easing quantization. 

\vspace{-1mm}
\noindent \textbf{Dual-Scale Quantizer.}  
To better leverage scaling flexibility, we formulate down-projection quantization as a dual-scale problem by embedding an additional (column-wise) scale directly into the quantization process, rather than applying it as static smoothing. We begin by revisiting standard per-channel quantization, as defined in Eqs.~\ref{eq:uniform_quant}–\ref{eq:uniform_dequant}, where each weight tensor is quantized using channel-wise scale and zero-point parameters. For simplicity, we abstract this process as a generic quantization operator $Q(\cdot)$, yielding $\widehat{\mathbf{W}} = Q(\mathbf{W})$. To refine the quantized weights, we introduce an additional column-wise scale factor $\mathbf{s}^c \in \mathbb{R}^{1 \times H}$, resulting in the dual-scale quantized form:
\begin{equation}
\widehat{\mathbf{W}} = Q(\mathbf{W}) \odot \mathbf{s}^c.
\end{equation}
Our objective is to minimize the reconstruction error between the original and quantized weights:
\begin{equation}
\min \left\| \mathbf{W} - \widehat{\mathbf{W}} \right\|_F^2 = \left\| \mathbf{W} - Q(\mathbf{W}) \odot \mathbf{s}^c \right\|_F^2.
\end{equation}
To efficiently solve this objective, we adopt an iterative optimization strategy. Specifically, we first freeze the quantization operator $Q(\cdot)$ and solve for the optimal $\mathbf{s}^c$ in closed form. Then, we fix $\mathbf{s}^c$ and update the quantized weights by applying $Q$ to the normalized weights $\mathbf{W} / \mathbf{s}^c$. This process is repeated until convergence. In practice, the iterative procedure converges within a few steps and effectively integrates the column-wise scale into standard quantization. After down-projection quantization, since the up-projection has already been quantized with per-channel scales, the additional column-wise factor $\mathbf{s}^c$ can be directly merged by multiplying it into the existing scales. This preserves inference equivalence without introducing any runtime overhead.
\vspace{-7mm}
\subsection{Deviation-Aware Correction}
\begin{figure}[t!]
    \centering
    \includegraphics[width=\linewidth]{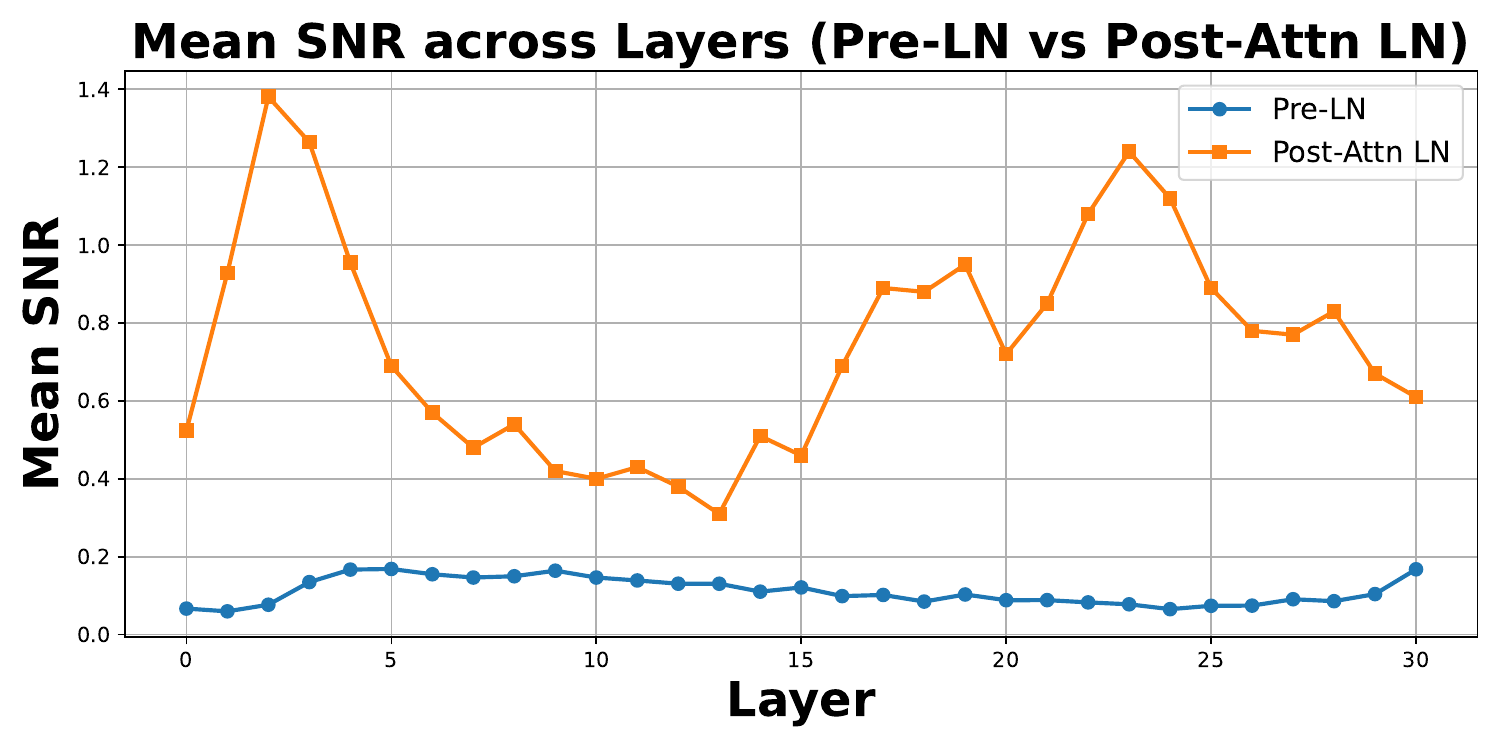}
    \vspace{-5mm}
    \caption{Mean SNR across transformer layers on LLaMA-3-8B. Post-attention LayerNorm exhibits consistently higher signal-to-noise ratios than Pre-LayerNorm, indicating stronger and more structured activation shifts caused by attention quantization.}
    \vspace{-7mm}
    \label{fig:fig4}
\vspace{-2mm}
\end{figure}

\vspace{-1mm}
\label{sec:DAC}
\noindent \textbf{Signal-to-Noise Ratio (SNR) Analysis.}
Based on our observations, weight-only quantization in transformer blocks induces activation shifts at subsequent LayerNorms. Quantizing attention causes a pronounced mean shift at the post-attention LayerNorm, while MLP quantization results in weaker, less structured deviations at the pre-LayerNorm of the next block. To quantify these effects, we introduce a signal-to-noise ratio (SNR) metric. Let $\mathbf{X} \in \mathbb{R}^{L \times H}$ denote the calibration input, where $L$ is the token count and $H$ the hidden dimension. Taking post-attention LayerNorm as an example, the full-precision output is:
\begin{equation}
    \mathbf{Y}_{fp} = \text{PostAttnLN}(\text{Attention}(\mathbf{X})),
\end{equation}
and the quantized counterpart as:
\begin{equation}
    \mathbf{Y}_{q} = \text{PostAttnLN}(Q(\text{Attention}(\mathbf{X}))).
\end{equation}
We define the activation deviation as:
\begin{equation}
    \Delta \mathbf{Y} = \mathbf{Y}_{fp} - \mathbf{Y}_q,
\end{equation}
which captures the activation shift at the post-attention LayerNorm caused by quantizing the attention module. Here, $\Delta \mathbf{Y} \in \mathbb{R}^{L \times H}$ contains the per-token deviations across $L$ tokens and $H$-dimensional features. The mean and variance of the deviation across tokens are computed as:
\begin{equation}
    \boldsymbol{\mu} = \frac{1}{L} \sum_{t=1}^{L} \Delta \mathbf{Y}^{(t)},
    \quad
    \boldsymbol{\sigma}^2 = \frac{1}{L} \sum_{t=1}^{L} \left( \Delta \mathbf{Y}^{(t)} - \boldsymbol{\mu} \right)^2,
\end{equation}
where all operations are element-wise over the feature dimension $H$, and broadcasting is applied as needed. The signal-to-noise ratio (SNR) is then defined as:
\begin{equation}
    \mathrm{SNR} = \frac{|\boldsymbol{\mu}|}{\boldsymbol{\sigma}^2}.
\end{equation}
A higher SNR indicates a consistent directional shift across tokens, while a lower SNR reflects unstructured or negligible deviation. Fig.~\ref{fig:fig4} shows the average SNR across all layers for both pre-LayerNorm and post-attention LayerNorm on LLaMA-3-8B. We observe that the SNR of the post-attention LayerNorm is consistently and significantly higher than that of the pre-LayerNorm, confirming that pronounced mean shifts indeed occur after attention quantization.

\vspace{-1mm}
\noindent \textbf{Bias Alignment for Post-Attention LayerNorm.}
Guided by the SNR analysis, we apply deviation correction only at post-attention LayerNorm layers, where the activation shifts are strong and consistent across tokens. For each such layer, we estimate a correction bias $\boldsymbol{\mu}$ by computing the mean deviation $\Delta \mathbf{Y}$ across a small calibration set. This bias is then added to the quantized output during inference:
\begin{equation}
\mathbf{Y}_{\text{aligned}} = \mathbf{Y}_q + \boldsymbol{\mu}.
\end{equation}
The correction term $\boldsymbol{\mu}$ is stored as an additional bias parameter inside the corresponding LayerNorm and is applied jointly during inference. Its parameter size and runtime overhead are negligible compared to the overall model.

\vspace{-1mm}
\noindent \textbf{Error Reduction via Deviation Correction.}  
To evaluate deviation correction, we analyze its impact on activation reconstruction error. The deviation between full-precision and quantized LayerNorm outputs is $\Delta \mathbf{Y} = \mathbf{Y}_{fp} - \mathbf{Y}_q$. For the $i$-th feature dimension, the expected squared deviation (MSE) across tokens can be decomposed as:
\begin{equation}
    \mathbb{E}[\|\Delta \mathbf{Y}_i\|^2] = \mu_i^2 + \sigma_i^2,
\end{equation}
where $\mu_i$ and $\sigma_i^2$ are the mean and variance of the deviation in the $i$-th channel, respectively. We apply deviation correction by adding a learned bias $\boldsymbol{\mu}$ to the quantized output, which shifts the deviation to:
\begin{equation}
    \Delta \mathbf{Y}_{\text{aligned}} = \mathbf{Y}_{fp} - \mathbf{Y}_{\text{aligned}} = \Delta \mathbf{Y} - \boldsymbol{\mu}.
\end{equation}
The expected squared error then becomes:
\begin{equation}
    \mathbb{E}[\|\Delta \mathbf{Y}_{\text{aligned},i}\|^2] = \sigma_i^2.
\end{equation}
Thus, the relative error reduction is:
\begin{equation}
    \frac{\mu_i^2 + \sigma_i^2 - \sigma_i^2}{\mu_i^2 + \sigma_i^2} = \frac{\mu_i^2}{\mu_i^2 + \sigma_i^2}.
\end{equation}
This ratio quantifies how much error is eliminated by correcting the mean shift. Notably, this is exactly the form of the signal-to-noise ratio (SNR)–based reduction:
\begin{equation}
    \frac{\mu_i^2}{\mu_i^2 + \sigma_i^2} = \frac{\mathrm{SNR}_i}{1 + \mathrm{SNR}_i},
    \quad \text{where} \quad \mathrm{SNR}_i = \frac{\mu_i^2}{\sigma_i^2}.
\end{equation}
Hence, dimensions with higher SNR benefit more from deviation correction, since the dominant error from mean shift can be effectively removed.

\vspace{-3mm}
\begin{algorithm}[h]
\caption{Main Framework of D$^2$Quant: inner details of each function are provided in the supplementary material.}
\label{alg1}
func $\operatorname{D^2Quant}$($\mathcal{M}$, $\mathcal{X}$)\\
{\bf Input:} $\mathcal{M}$ - Pre-trained model with $L$ blocks \\
\hspace*{0.43in}$ \mathcal{X}$ - Calibration data \\
{\bf Output:} $\widehat{\mathcal{M}}$ - Quantized model
\begin{algorithmic}[1]
\STATE $\widehat{\mathcal{M}} \leftarrow \mathcal{M}$ \mycomment{Initialize quantized model}
\FOR{$l=1$ to $L$}
    \STATE $X_l \leftarrow \mathrm{GetPostAttnLNAct}(\mathcal{M}_l,\mathcal{X})$
    \FOR{$W \in \{W_q^l,W_k^l,W_v^l,W_o^l\}$}
        \STATE $W \leftarrow \mathrm{Quantizer}(W)$
    \ENDFOR
    \STATE $\widehat{\mathcal{M}}_l \leftarrow \mathrm{WriteBack}(W_q^l,W_k^l,W_v^l,W_o^l)$
    \STATE $\widehat{X}_l \leftarrow \mathrm{GetPostAttnLNAct}(\widehat{\mathcal{M}}_l,\mathcal{X})$
    \STATE $\mathrm{PostAttnLN}_l \leftarrow \mathrm{DAC}(\mathrm{PostAttnLN}_l, X_l, \widehat{X}_l)$
    \STATE $\widehat{\mathcal{M}}_l \leftarrow \mathrm{WriteBack}(\mathrm{PostAttnLN}_l)$
    \FOR{$W \in \{W_{up}^l,W_{gate}^l\}$}
        \STATE $W \leftarrow \mathrm{Quantizer}(W)$
    \ENDFOR
    \STATE $W_{down}^l, s_c\leftarrow \mathrm{DSQ}(W_{down}^l)$
    \STATE $W_{up}^l \leftarrow \mathrm{MergeScale}(W_{up}^l,s_c) $
    \STATE $\widehat{\mathcal{M}}_l \leftarrow \mathrm{WriteBack}(W_{up}^l,W_{gate}^l,W_{down}^l) $ 
\STATE $\mathcal{X}\leftarrow\mathrm{Forward}(\widehat{\mathcal{M}}_l,\mathcal{X})$\mycomment{Update calibration data for the next block}
\ENDFOR
\STATE \textbf{return} $\widehat{\mathcal{M}}$
\end{algorithmic}
\end{algorithm}
\vspace{-3mm}

\vspace{-2mm}
\subsection{D$^2$Quant Pipeline}
\vspace{-1.5mm}
\label{sec:pipeline}
As shown in Algorithm~\ref{alg1}, D$^2$Quant follows a block-wise weight-only PTQ pipeline that integrates DSQ and DAC into a unified framework. Starting from a pre-trained model, we iterate over blocks and first collect the full-precision post-attention LayerNorm activations as the alignment target. We then quantize the attention module and recompute the corresponding activations to drive DAC. DAC calibrates the post-attention LayerNorm to mitigate activation drift caused by attention quantization. Next, we quantize the FFN up/gate projections and apply DSQ to the down-projection to derive a column-wise scale, which is folded into the quantized up-projection for deployment-friendly inference. Finally, we forward the updated quantized block to refresh the calibration data for the next block, yielding an end-to-end D$^2$Quant pipeline that leverages DAC for activation alignment and DSQ for accurate down-projection quantization.

\begin{table*}[!ht]
\setlength{\tabcolsep}{5pt}
\small
\centering
\caption{
\textbf{Performance of 2-bit weight quantization on the Qwen-3 series.} We report perplexity ($\downarrow$) on WikiText2 and C4, accuracy ($\uparrow$) on MMLU, and on seven commonsense tasks with their average (Avg.). $^{\dagger}$ denotes methods with QuaRot rotation. Best results are in \textbf{bold}. 
}
\label{tab:tab1}
\vspace{-1mm}
\begin{adjustbox}{max width=\linewidth}
\begin{tabular}{c|l|cc|c|cccccccc}
\hline
\rowcolor{color3}
\textbf{Model} & \textbf{Method}  &
\textbf{Wiki2$(\downarrow)$} & \textbf{C4$(\downarrow)$} & \textbf{MMLU$(\uparrow)$} &
\textbf{PiQA} & \textbf{Hella.} & \textbf{Arc-E} & \textbf{Arc-C} &
 \textbf{Wino.} & \textbf{RTE} & \textbf{OBQA} &
\textbf{Avg.$(\uparrow)$} \\
\hline
 & \rule{0pt}{2.0ex}FP16 & 9.72 & 15.43 & 72.96 & 77.64 & 74.90 & 80.81 & 57.00 & 68.03 & 77.98 & 41.80 & 68.31 \\
\cdashline{2-13}
\multirow{6}{*}{\shortstack{Qwen3\\8B}} 
      & \rule{0pt}{2.0ex}GPTQ & 23.28 & 55.55 & 37.05 & 65.23 & 47.27 & 48.11 & 31.14 & 55.56 & 57.04 & 30.40 & 47.82 \\
      & GPTQ$^{\dagger}$ & 15.27 & 28.60 & 46.53 & 66.92 & 53.03 & 56.52 & 35.92 & 61.88 & 71.48 & 32.60 & 54.05 \\
    & GPTAQ & 20.24 & 52.02 & 30.37 & 61.81 & 42.05 & 45.88 & 28.67 & 58.33 & 60.29 & 29.60 & 46.66 \\
      & GPTAQ$^{\dagger}$ & 15.61 & 29.91 & 43.49 & 66.81 & 49.42 & 54.80 & 33.02 & 61.64 & 64.26 & 31.40 & 51.62 \\
      & BoA & 17.67 & 40.37  & 40.37 & 65.45  & 46.93  & 55.18 & 32.34  & 59.75  & 54.51 & 31.2  & 49.34  \\
      & D$^2$Quant & \textbf{14.10} & \textbf{25.96} & \textbf{49.94} & \textbf{70.51} & \textbf{55.76} & \textbf{66.84} & \textbf{39.25} & \textbf{61.96} & \textbf{71.84} & \textbf{34.40} & \textbf{57.22} \\
\hline
 & \rule{0pt}{2.0ex}FP16 & 8.64 & 13.82 & 77.11 & 79.76 & 78.89 & 83.04 & 60.24 & 73.01 & 77.62 & 46.80 & 71.34 \\
\cdashline{2-13}
\multirow{6}{*}{\shortstack{Qwen3\\14B}} 
       & \rule{0pt}{2.0ex}GPTQ & 12.71 & 28.68 & 53.20 & 71.93 & \textbf{61.87} & 62.84 & 39.08 & 64.72 & 49.46 & 36.40 & 55.19 \\
       & GPTQ$^{\dagger}$ & 12.92 & 38.23 & 46.90 & 71.98 & 54.43 & 67.30 & 40.78 & 66.30 & 66.43 & 37.40 & 57.80 \\
     & GPTAQ & 12.91 & 28.19 & 51.33 & 71.06 & 60.46 & 66.58 & 41.47 & 66.77 & 61.73 & 39.00 & 58.15 \\
       & GPTAQ$^{\dagger}$ & 12.48 & 23.93 & 56.12 & 71.55 & 58.85 & 64.73 & 40.36 & 65.90 & \textbf{81.59} & 35.00 & 59.71 \\
        & BoA & 14.10 & 29.33 & 44.95  & 70.24 & 54.25 & 58.04 & 35.49 & 65.51 & 74.01 & 32.80 & 55.76\\
       & D$^2$Quant & \textbf{11.88} & \textbf{22.40} & \textbf{58.36} & \textbf{72.31} & 60.98 & \textbf{71.51} & \textbf{44.54} & \textbf{66.85} & 74.73 & \textbf{39.20} & \textbf{61.45} \\
\hline
 & \rule{0pt}{2.0ex}FP16 & 7.61 & 12.45 & 80.76 & 82.05 & 82.62 & 83.21 & 61.26 & 73.01 & 76.17 & 46.00 & 72.05 \\
\cdashline{2-13}
\multirow{6}{*}{\shortstack{Qwen3\\32B}} 
       & \rule{0pt}{2.0ex}GPTQ & 11.13 & 25.27 & 63.04 & 73.78 & 68.76 & 64.06 & 44.71 & 68.75 & \textbf{74.01} & 40.20 & 62.04 \\
       & GPTQ$^{\dagger}$ & 10.61 & 19.38 & 63.03 & 72.52 & 68.80 & 67.34 & 48.12 & 67.88 & 67.87 & \textbf{41.60} & 62.02 \\
     & GPTAQ & 10.78 & 24.12 & 62.54 & 73.94 & 68.33 & 66.84 & 44.62 & 69.06 & 67.51 & 38.20 & 61.21 \\
       & GPTAQ$^{\dagger}$ & 10.44 & 20.03 & 62.61 & 74.32 & 68.50 & 67.85 & 46.33 & 68.51 & 69.31 & 39.80 & 62.09 \\
        & BoA & N/A & N/A & N/A & N/A & N/A & N/A & N/A & N/A & N/A & N/A & N/A \\
       & D$^2$Quant & \textbf{9.71} & \textbf{17.10} & \textbf{64.31} & \textbf{76.12} & \textbf{70.47} & \textbf{75.51} & \textbf{51.11} & \textbf{70.09} & 68.23 & 40.40 & \textbf{64.56} \\
\hline
\end{tabular}
\end{adjustbox}
\label{main_results}
\vspace{-3mm}
\end{table*}

\newpage
\section{Experiments}
\vspace{-1mm}
\subsection{Settings}
\vspace{-1mm}
\noindent \textbf{Implementation Details.}
All experiments are conducted using PyTorch~\citep{paszke2019pytorch} and the HuggingFace Transformers library~\citep{wolf-etal-2020-transformers} on NVIDIA A800-80GB GPUs. Except for benchmark evaluations on 70B-scale models, which are performed using three GPUs, all quantization and evaluation experiments are conducted on a single GPU. We use 128 samples from the WikiText-2 dataset~\citep{merity_pointer_2016} with a sequence length of 2048 as the calibration set during quantization, and all quantized models adopt a fixed quantization block size of 128. We implement 15 iterations for Dual-Scale Quantizer (DSQ) to ensure the convergence of quantization parameters.

\vspace{-1mm}
\noindent\textbf{Baselines.} We compare our method against GPTQ~\citep{frantar_gptq_2023} and GPTAQ~\citep{li_gptaq_2025}, two representative weight-only PTQ approaches for LLMs. In addition, we implement the randomized Hadamard transform proposed in Quarot~\citep{ashkboos_quarot_2024} as a weight pre-processing step on top of these baselines to smooth the weight distributions. We denote the resulting variants as GPTQ$^{\dagger}$ (GPTQ+Quarot) and GPTAQ$^{\dagger}$ (GPTAQ+Quarot). We further compare with BoA~\citep{kim_boa_nodate}, a recent PTQ method that uses more accurate Hessian estimation.

\vspace{-1mm}
\noindent \textbf{Models and Evaluation.} We evaluate our method on a range of pre-trained LLMs, including LLaMA-3 (8B/70B)~\citep{grattafiori_llama_2024}, LLaMA-3.1 (8B/70B), and Qwen-3 (8B/14B/32B)~\citep{qwen3technicalreport}. We assess quantized models using both language modeling and downstream benchmarks. We report perplexity on WikiText2~\citep{merity_pointer_2016} and C4~\citep{2020t5} with a sequence length of 2048 tokens, and measure zero-shot accuracy on PIQA~\citep{bisk_piqa_2019}, HellaSwag~\citep{zellers_hellaswag_2019}, ARC-Easy/Challenge~\citep{clark_think_2018}, WinoGrande~\citep{sakaguchi_winogrande_2019}, RTE~\citep{chakrabarty2021figurative}, and OpenBookQA~\citep{mihaylov_can_2018}. We evaluate on MMLU~\citep{hendrycks_measuring_2021}, a multi-domain benchmark for knowledge-intensive reasoning.

\begin{table*}[h]
\setlength{\tabcolsep}{5pt}
\small
\centering
\caption{
\textbf{2-bit weight quantization results on the LLaMA-3/3.1 series.}
Perplexity ($\downarrow$) on WikiText2/C4 and accuracy ($\uparrow$) on MMLU and seven commonsense tasks performance with their average (Avg.). $^{\dagger}$ denotes methods with QuaRot rotation. Best results are in \textbf{bold}.
}
\label{tab:tab2}
\vspace{-1mm}
\begin{adjustbox}{max width=\linewidth}
\begin{tabular}{c|l|cc|c|cccccccc}
\hline
\rowcolor{color3}
\textbf{Model} & \textbf{Method}  &
\textbf{Wiki2$(\downarrow)$} & \textbf{C4$(\downarrow)$} & \textbf{MMLU$(\uparrow)$} &
\textbf{PiQA} & \textbf{Hella.} & \textbf{Arc-E} & \textbf{Arc-C} &
 \textbf{Wino.} & \textbf{RTE} & \textbf{OBQA} &
\textbf{Avg.$(\uparrow)$} \\
\hline
& FP16 & \rule{0pt}{2.0ex}6.14 & 9.44 & 62.16 & 80.69  & 79.17  & 77.86 & 53.16 & 73.32  & 67.87 & 45.00 & 68.15 \\
\cdashline{2-13}
\multirow{6}{*}{\shortstack{LLaMA3\\8B}} 
      & \rule{0pt}{2.0ex}GPTQ & 17.33 & 67.14 & 23.27 & 54.79 & 42.95 & 31.31 & 22.18 & 54.14 & 52.71 & 30.20 & 41.18 \\
      & GPTQ$^{\dagger}$ & 28.97 & 79.97 & 23.10 & 55.82 & 34.74 & 34.55 & 21.33 & 53.51 & 52.71 & 24.80 & 39.64 \\
    & GPTAQ & 14.17 & 132.13 & 23.02 & 57.67 & 40.41 & 35.23 & 23.29 & 56.35 & 52.35 & 27.00 & 41.76 \\
      & GPTAQ$^{\dagger}$ & 14.28 & 35.91 & 24.46 & \textbf{62.68} & 47.36 & \textbf{45.58} & 27.99 & 57.62 & \textbf{54.87} & 30.20 & \textbf{46.61} \\
      & BoA & 23.13  & 66.96  & 22.95  & 58.11 & 39.15 & 39.27 & 25.00  & 55.33 & 52.71 & 28.8 & 40.16  \\
      & D$^2$Quant & \textbf{11.88} & \textbf{34.62} & \textbf{30.02} & 60.23 & \textbf{50.77} & 42.13 & \textbf{27.99} & \textbf{60.69} & 52.71 & \textbf{30.60} & 46.45 \\
\hline
 & \rule{0pt}{2.0ex}FP16 & 2.86  & 7.17  & 75.15  & 84.44 & 84.97 & 86.15 & 64.51 & 80.58 & 68.59 & 48.40 & 73.95  \\
\cdashline{2-13}
\multirow{6}{*}{\shortstack{LLaMA3\\70B}} 
        & \rule{0pt}{2.0ex}GPTQ  & 9.31 & 53.63 
 & 25.23 & \textbf{74.65} & 46.80 & 63.59 & 39.25 & 57.62 & 53.07 & 29.00 & 49.53   \\
        & GPTQ$^{\dagger}$ & 19.76  & 49.56  & 23.34  & 52.07 & 42.22 & 29.88 & 21.42 & 54.93 & 52.71 & 31.6 & 39.52 \\
      & GPTAQ & 9.18  & 27.26  & \textbf{40.26}  & 74.48 & \textbf{62.00} & 62.42 & 36.69 & 61.25 & \textbf{56.32} & 32.4  & 55.08   \\
        & GPTAQ$^{\dagger}$ & 15.97   & 41.45  & 24.78   & 56.37 & 44.05 & 32.53 & 20.39 & 55.72 & 52.71 & 29.6 & 40.98  \\
        & BoA & N/A & N/A & N/A & N/A & N/A & N/A & N/A & N/A & N/A & N/A & N/A \\
        & D$^{2}$Quant &\textbf{9.04}    & \textbf{24.92}  & 32.95   & 72.31 &56.82 &\textbf{64.60}  &\textbf{40.10} &\textbf{66.69} &53.43&\textbf{37.4}& \textbf{55.91} \\

\hline
 & \rule{0pt}{2.0ex}FP16 & 6.24 & 9.54 & 63.32 & 81.07 & 78.90 & 81.27 & 53.41 & 73.80 & 70.40 & 44.80 & 69.09 \\
\cdashline{2-13}
\multirow{6}{*}{\shortstack{LLaMA3.1\\8B}} 
        & \rule{0pt}{2.0ex}GPTQ & 18.61 & 120.14 & 23.76 & 59.41 & 44.77 & 42.13 & 25.68 & 53.75 & 52.35 & 30.20 & 44.04 \\
        & GPTQ$^{\dagger}$ & 24.60 & 77.78 & 23.17 & 60.66 & 34.98 & 38.68 & 22.95 & 53.59 & 52.35 & 25.80 & 41.29 \\
     & GPTAQ & 14.42 & 53.12 & 22.95 & 59.47 & 39.73 & 43.73 & 27.13 & 54.46 & 52.71 & 30.20 & 43.92 \\
        & GPTAQ$^{\dagger}$ & 13.61 & 32.92 & 24.58 & 64.20 & 46.47 & 47.43 & 27.90 & 56.43 & 53.43 & 30.00 & 46.55 \\
      & BoA & 23.27  & 61.30  & 23.01  & 61.21 & 41.36 & 41.67 & 24.57 & 55.41 & 52.35 & 26.4 & 43.28 \\
        & D$^2$Quant & \textbf{11.54} & \textbf{27.19} & \textbf{28.85} & \textbf{69.26} & \textbf{53.73} & \textbf{52.48} &\textbf{32.25} & \textbf{60.85} & \textbf{56.32} & \textbf{32.40} & \textbf{51.04} \\
\hline
 & \rule{0pt}{2.0ex}FP16 & 2.81  & 7.11  & 75.31  & 84.28  & 85.00 & 86.53 & 64.85 & 79.24 & 70.04 & 48.00 & 73.99  \\
\cdashline{2-13}
\multirow{6}{*}{\shortstack{LLaMA3.1\\70B}} 
        & \rule{0pt}{2.0ex}GPTQ & 12.09  & 248.64  & 38.11 & 71.65 & 56.48 & \textbf{62.92} & 38.40 & 61.09 & \textbf{57.04} & 32.60 & 53.25  \\
        & GPTQ$^{\dagger}$ & 14.00  & 34.39  & 26.45  & 60.83 & 55.42 & 43.10 & 24.83 & 58.80 & 51.26 & 28.00 & 45.32  \\
  & GPTAQ & 16.38  & 72.04  & 23.03  & 62.35 & 43.58 & 45.12 & 27.13 & 53.99 & 54.51 & 28.40 & 42.62  \\
        & GPTAQ$^{\dagger}$ & 12.53  & 28.52  & 30.19  & 64.09 & 50.00 & 42.47 & 24.91 & 57.30 & 54.15 & 31.40 & 46.24  \\
        & BoA & N/A & N/A & N/A & N/A & N/A & N/A & N/A & N/A & N/A & N/A & N/A \\
        & D$^{2}$Quant &\textbf{8.63}   &\textbf{17.74}   &\textbf{40.46}     & \textbf{73.94}& \textbf{61.87} & 62.58&\textbf{39.93} &\textbf{69.61}&53.79&\textbf{38.00}&\textbf{57.10}  \\
\hline
\end{tabular}
\end{adjustbox}
\label{main_results}
\vspace{-3mm}
\end{table*}

\vspace{-2mm}
\subsection{Main Results}
\vspace{-1mm}
\noindent \textbf{Perplexity Evaluation.}
First, we evaluate the language modeling performance of D$^2$Quant under 2-bit weight quantization. Tab.~\ref{tab:tab1} and Tab.~\ref{tab:tab2} report perplexity on WikiText2 and C4 across different model families. Compared to baselines including GPTQ, GPTQ$^{\dagger}$, GPTAQ, GPTAQ$^{\dagger}$, and BoA, D$^2$Quant consistently achieves lower perplexity at all scales. On Qwen-3-8B, it reduces WikiText2 perplexity from 20.24 (GPTAQ) to 14.10 and C4 from 52.02 to 25.96. On Qwen-3-14B, it further lowers C4 perplexity to 22.40 versus 28.19 with GPTAQ. For Qwen-3-32B, D$^2$Quant achieves 9.71 (WikiText2) and 17.10 (C4), outperforming all baselines by a large margin. Similar trends hold for LLaMA models: On LLaMA-3.1-8B, D$^2$Quant obtains 11.54 on WikiText2 and 27.19 on C4, substantially better than GPTQ (18.61 / 120.14) and GPTAQ (14.42 / 53.12). These consistent gains across diverse model families clearly demonstrate the robustness and effectiveness of D$^2$Quant in preserving language modeling quality under aggressive 2-bit quantization.

\vspace{-1mm}
\noindent \textbf{Zero-Shot Accuracy Evaluation.}
We further evaluate D$^2$Quant on seven representative zero-shot reasoning benchmarks. As shown in Tab.~\ref{tab:tab1} and Tab.~\ref{tab:tab2}, D$^2$Quant outperforms existing baselines across most tasks and model sizes. On the Qwen series, for example, D$^2$Quant raises the average accuracy of Qwen-3-32B from 61.21 (GPTAQ) to 64.56, achieving a +3.35 gain. On LLaMA-3.1-8B, the improvement is even more pronounced, with average accuracy rising from 44.04 (GPTQ) to 51.04, a +6.99 gain. These consistent improvements clearly demonstrate the strong ability of D$^2$Quant to retain robust reasoning and commonsense capabilities under aggressive 2-bit quantization.

\vspace{-0.7mm}
\noindent \textbf{MMLU Evaluation.}
To further assess the reasoning and knowledge retention of quantized models, we evaluate D$^2$Quant on the MMLU benchmark, with results summarized in Tab.~\ref{tab:tab1} and Tab.~\ref{tab:tab2}. D$^2$Quant improves MMLU accuracy across nearly all model scales and architectures. On Qwen-3-32B, it achieves a gain of 1.27 points over GPTQ, while on LLaMA-3.1-8B the improvement reaches 5.09 points. These results indicate that D$^2$Quant is highly effective at mitigating knowledge degradation from aggressive 2-bit weight quantization, leading to stronger factual and reasoning performance in general-purpose language tasks without requiring any task-specific adaptation.

\vspace{-0.7mm}
\noindent \textbf{Additional Experimental Results.}
Due to space limits, we include additional results in the supplementary material, including 3-bit weight-only evaluations, further demonstrating the flexibility and robustness of D$^2$Quant across bit-widths. 

% \newpage
\vspace{-1mm}
\subsection{Ablation Study}
\vspace{-1mm}
\noindent \textbf{Effect of DSQ and DAC Components.}
Table~\ref{tab:ablaiton_dsq_dac} reports a component-wise ablation on Qwen-3-8B under 2-bit quantization. We evaluate two perplexity metrics (WikiText2 and C4), the MMLU benchmark, and an average accuracy score across seven zero-shot classification tasks (Acc). Introducing Dual-Scale Quantizer (DSQ) yields consistent gains across all metrics, notably improving MMLU by +3.48 and average accuracy by +3.27. Deviation-Aware Correction (DAC) also provides consistent gains, with notable improvements in downstream accuracy. When combined, our D$^2$Quant (DSQ+DAC) achieves the best results across all benchmarks, validating their complementary strengths.

\vspace{-1mm}
\begin{table}[h]
\setlength{\tabcolsep}{5pt}
\small
\centering
\label{ablation1}
\vspace{-2mm}
\caption{
\textbf{Effect of DSQ and DAC Components.}
Component-wise breakdown analysis of DSQ and DAC on Qwen-3-8B (2-bit).}
\label{tab:ablaiton_dsq_dac}
\vspace{-1mm}
\begin{adjustbox}{width=1\columnwidth}
\begin{tabular}{c|l|cc|c|c}
\hline
\rowcolor{color3}
\textbf{Model} & \textbf{Method}  &
\textbf{Wiki2$(\downarrow)$} & \textbf{C4$(\downarrow)$} & \textbf{MMLU$(\uparrow)$} &\textbf{Acc$(\uparrow)$} \\
\hline
\multirow{4}{*}{Q3-8B}
 &\rule{0pt}{2.0ex}Baseline & 14.72 & 27.47 & 45.49 & 53.94 \\
\cdashline{2-6}
&\rule{0pt}{2.0ex}+DSQ     & 14.41 & 26.95 & 48.97 & 57.21 \\
&+DAC     & 14.63 & 27.00 & 45.82 & 54.63 \\
&\textbf{+DSQ+DAC} & \textbf{14.10} & \textbf{25.96} & \textbf{49.94} & \textbf{57.22} \\
\cdashline{2-6}
\hline
\end{tabular}
\end{adjustbox}
\vspace{-1.2mm}
\end{table}

\vspace{-1.7mm}
\noindent \textbf{Impact of Calibration Set Size for DAC.}
Table~\ref{tab:dac_calib_size} presents the performance of DAC with varying calibration set sizes on Qwen-3-8B using 2-bit quantization. As expected, a larger calibration set leads to more accurate bias correction and better performance. The results show a steady improvement as the calibration size increases from 16 to 128, with the best performance achieved at 128 samples, where both Wiki2 and C4 perplexity are minimized, and MMLU accuracy reaches 49.94. However, when the calibration set size is too small (e.g., 16), the bias correction becomes less accurate, leading to a performance drop, particularly in downstream accuracy. Based on these observations, we choose 128 as the optimal calibration set size for DAC, as it balances computational efficiency and correction accuracy.

\vspace{-1.5mm}
\begin{table}[t]
\setlength{\tabcolsep}{5pt}
\small
\centering
\caption{
\textbf{Impact of Calibration Set Size for DAC.}
DAC performance with varying calibration sizes on Qwen-3-8B (2-bit).
}
\label{tab:dac_calib_size}
\vspace{-1mm}
\begin{adjustbox}{width=1\columnwidth}
\begin{tabular}{c|l|cc|c|c}
\hline
\rowcolor{color3}
\textbf{Model} & \textbf{Method (Cal. Size)} &
\textbf{Wiki2$\downarrow$} & \textbf{C4$\downarrow$} & \textbf{MMLU$\uparrow$} & \textbf{Acc$\uparrow$} \\
\hline
\multirow{6}{*}{Q3-8B}
 &\rule{0pt}{2.0ex}Baseline & 14.72 & 27.47 & 45.49 & 53.94 \\
\cdashline{2-6}
& \rule{0pt}{2.0ex}+DAC (16)  & 15.75 & 31.15 & 42.06 & 52.51 \\
& +DAC (32)  & 14.86 & 27.40 & 49.44 & 56.07 \\
& +DAC (64)  & 14.71 & 26.83 & 48.39 & \textbf{57.80} \\
& +DAC (128) & \textbf{14.10} & \textbf{25.96} & \textbf{49.94} & 57.22 \\
& +DAC (256) & 14.19 & 26.09 & 49.67 & 57.80 \\
\hline
\end{tabular}
\end{adjustbox}
\vspace{-6mm}
\end{table}

\vspace{1mm}
\noindent \textbf{Design Ablation for DSQ.}
Table~\ref{tab:dsq_ablation} presents a design ablation of DSQ on Qwen-3-8B under 2-bit quantization. Static smoothing applied to down-projection weights does not consistently improve performance and can slightly degrade both perplexity and accuracy, suggesting that fixed rescaling is insufficient to address quantization distortions. In contrast, DSQ dynamically incorporates column-wise scaling into the quantization objective, yielding consistent improvements across tasks. The results also highlight the importance of iterative refinement: performance improves steadily with more iterations and saturates around 15, indicating convergence. We thus adopt 15 iterations as the default setting, balancing effectiveness and overall efficiency.

\vspace{-3mm}
\begin{figure}[h]
    \centering
    \includegraphics[width=\linewidth]{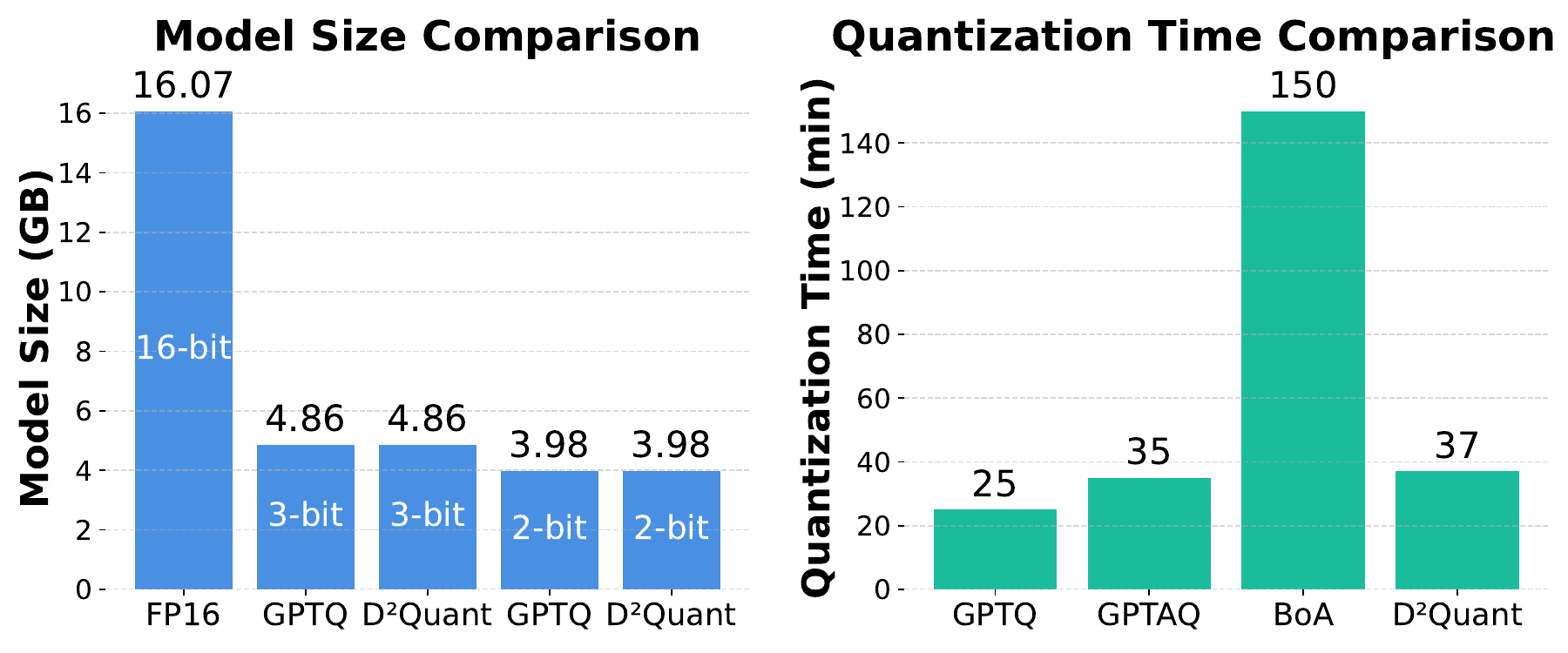}
    \vspace{-7mm}
    \caption{Model size and quantization time on LLaMA-3-8B.}
    \vspace{-4mm}
    \label{fig:fig5}
\end{figure}
\vspace{-1mm}

% \vspace{-1mm}
\subsection{Time and Memory Analyses}
\vspace{-0.5mm}
As shown in Fig.~\ref{fig:fig5}, D$^2$Quant exhibits favorable efficiency in both model size and quantization time. In terms of model size, our method achieves an almost identical footprint to GPTQ under the same bit-width, since the additional parameters introduced by adding bias correction in the LayerNorm are negligible compared to the overall model size. Regarding quantization time, D$^2$Quant incurs a modest overhead relative to GPTQ and GPTAQ, mainly due to the LayerNorm updates and the iterative DSQ optimization on the down-projection. Nevertheless, it remains faster than BoA, which relies on expensive Hessian-based optimization, demonstrating a strong balance between performance and efficiency. Thus, D$^2$Quant enables scalable LLM deployment.

\vspace{-0.5mm}
\section{Discussions and Future Works}
\vspace{-1mm}
\noindent \textbf{Broader Applicability of DSQ.}
This work applies the Dual-Scale Quantizer (DSQ) to down-projection by leveraging column scale absorption into the preceding up-projection. The same idea may extend to QKV and up/gate projections, where column scales can be folded into the preceding LayerNorm. However, this requires all branches (e.g., Q, K, V) to share a column scale, posing design constraints. Extending DSQ under these constraints is a promising direction.

\vspace{-1mm}
\noindent \textbf{Identifying and Correcting More Activation Shift Patterns.}
This work highlights a consistent mean shift at the post-attention LayerNorm caused by quantizing the attention module. However, weight-only quantization may induce other types of activation distribution shifts in different parts of the model. Identifying such patterns and designing general correction mechanisms beyond mean bias could improve the robustness and generalization of quantized models, making this a valuable direction for future work.

\vspace{-1.5mm}
\begin{table}[t]
\setlength{\tabcolsep}{5pt}
\small
\centering
\caption{
\textbf{Design Ablation Study for DSQ.}
Comparison of static smoothing and DSQ at different iterations on Qwen-3-8B (2-bit).
}
\label{tab:dsq_ablation}
\vspace{-1mm}
\begin{adjustbox}{width=1\columnwidth}
\begin{tabular}{c|l|cc|c|c}
\hline
\rowcolor{color3}
\textbf{Model} & \textbf{Method} &
\textbf{Wiki2$\downarrow$} & \textbf{C4$\downarrow$} & \textbf{MMLU$\uparrow$} & \textbf{Acc$\uparrow$} \\
\hline
\multirow{6}{*}{Q3-8B}
 & \rule{0pt}{2.27ex}Baseline          & 14.72 & 27.47 & 45.49 & 53.94 \\
\cdashline{2-6}
 & \rule{0pt}{2.27ex}+Static Smooth     & 14.85 & 27.72 & 44.99 & 52.28 \\
 & +DSQ (iterations=0)        &  15.73    &  31.90    &  46.05    &  53.71    \\
 & +DSQ (iterations=1)        &  14.90    &  28.62    &  48.70    &  57.01    \\
 & +DSQ (iterations=3)        &  14.66    &  26.98    &  48.76   &  57.15    \\
 & +DSQ (iterations=15)       & \textbf{14.41} & \textbf{26.95} & \textbf{48.97} & \textbf{57.21} \\
\hline
\end{tabular}
\end{adjustbox}
\vspace{-6mm}
\end{table}

\vspace{-2mm}
\section{Conclusion}
\vspace{-1.5mm}
In this work, we revisit weight-only PTQ for LLMs at sub-4-bit precision and make two key observations: (1) a smoothing-equivalent transformation between the up- and down-projection significantly eases down-projection quantization without affecting the up-projection; (2) attention module quantization induces a pronounced mean shift at the post-attention LayerNorm. Building on these insights, we propose D$^2$Quant, a unified weight-only PTQ framework that improves sub-4-bit quantization performance. On the weight side, the proposed Dual-Scale Quantizer (DSQ) improves the robustness of down-projection quantization without increasing the bit budget or inference overhead. On the activation side, Deviation-Aware Correction (DAC) mitigates quantization-induced activation shifts in post-attention LayerNorm. Extensive experiments across multiple LLM families and evaluation benchmarks demonstrate that D$^2$Quant consistently outperforms prior SOTA weight-only PTQ methods in the sub-4-bit regime. Importantly, D$^2$Quant is composed of structurally simple components, many of which are absorbable, making the framework easy to integrate into mainstream inference pipelines. More broadly, this work suggests that effective weight-only PTQ requires jointly preserving critical weight structures and correcting quantization-induced activation shifts. We hope this perspective offers useful insights for developing more robust and deployable low-bit quantization methods.

% \vspace{-1mm}
% \section*{Impact Statement}
% \vspace{-1mm}
% This paper presents work whose goal is to advance the field of Machine Learning. There are many potential societal consequences of our work, none of which we feel must be specifically highlighted here.


\bibliography{icml2026}

@inproceedings{zhao2024atom,
  title={Atom: Low-bit quantization for efficient and accurate llm serving},
  author={Zhao, Yilong and Lin, Chien-Yu and Zhu, Kan and Ye, Zihao and Chen, Lequn and Zheng, Size and Ceze, Luis and Krishnamurthy, Arvind and Chen, Tianqi and Kasikci, Baris},
  booktitle={MLSys},
  year={2024}
}

@inproceedings{huang2024billm,
  title={Billm: Pushing the limit of post-training quantization for llms},
  author={Huang, Wei and Liu, Yangdong and Qin, Haotong and Li, Ying and Zhang, Shiming and Liu, Xianglong and Magno, Michele and Qi, Xiaojuan},
  booktitle={ICML},
  year={2024}
}

@inproceedings{wu2025quantcache,
  title={Quantcache: Adaptive importance-guided quantization with hierarchical latent and layer caching for video generation},
  author={Wu, Junyi and Li, Zhiteng and Hui, Zheng and Zhang, Yulun and Kong, Linghe and Yang, Xiaokang},
  booktitle={ICCV},
  year={2025}
}

@inproceedings{li2024arb,
  title={Arb-llm: Alternating refined binarizations for large language models},
  author={Li, Zhiteng and Yan, Xianglong and Zhang, Tianao and Qin, Haotong and Xie, Dong and Tian, Jiang and Kong, Linghe and Zhang, Yulun and Yang, Xiaokang and others},
  booktitle={ICLR},
  year={2025}
}

@inproceedings{zhang2025quant,
  title={Quant-dLLM: Post-Training Extreme Low-Bit Quantization for Diffusion Large Language Models},
  author={Zhang, Tianao and Li, Zhiteng and Yan, Xianglong and Qin, Haotong and Guo, Yong and Zhang, Yulun},
  booktitle={ICLR},
  year={2026}
}

@inproceedings{yan2025pt,
title={PT$^2$-LLM: Post-Training Ternarization for Large Language Models}, 
  author={Yan, Xianglong and Bao, Chengzhu and Li, Zhiteng and Zhang, Tianao and Yang, Kaicheng and Qin, Haotong and Xie, Ruobing and Sun, Xingwu and Zhang, Yulun},
  booktitle={ICLR},
  year={2026}
}

@article{yan2025progressive,
  title={Progressive binarization with semi-structured pruning for llms},
  author={Yan, Xianglong and Zhang, Tianao and Li, Zhiteng and Qin, Haotong and Zhang, Yulun},
  journal={arXiv preprint arXiv:2502.01705},
  year={2025}
}

@inproceedings{lin2025qserve,
  title={Qserve: W4a8kv4 quantization and system co-design for efficient llm serving},
  author={Lin, Yujun and Tang, Haotian and Yang, Shang and Zhang, Zhekai and Xiao, Guangxuan and Gan, Chuang and Han, Song},
  booktitle={MLSys},
  year={2025}
}

@inproceedings{dettmers2022gpt3,
  title={Gpt3. int8 (): 8-bit matrix multiplication for transformers at scale},
  author={Dettmers, Tim and Lewis, Mike and Belkada, Younes and Zettlemoyer, Luke},
  booktitle={NeurIPS},
  year={2022}
}

@inproceedings{lin2024duquant,
  title={Duquant: Distributing outliers via dual transformation makes stronger quantized llms},
  author={Lin, Haokun and Xu, Haobo and Wu, Yichen and Cui, Jingzhi and Zhang, Yingtao and Mou, Linzhan and Song, Linqi and Sun, Zhenan and Wei, Ying},
  booktitle={NeurIPS},
  year={2024}
}

@inproceedings{hu2025ostquant,
  title={Ostquant: Refining large language model quantization with orthogonal and scaling transformations for better distribution fitting},
  author={Hu, Xing and Cheng, Yuan and Yang, Dawei and Xu, Zukang and Yuan, Zhihang and Yu, Jiangyong and Xu, Chen and Jiang, Zhe and Zhou, Sifan},
  booktitle={ICLR},
  year={2025}
}

@inproceedings{sun2024flatquant,
  title={Flatquant: Flatness matters for llm quantization},
  author={Sun, Yuxuan and Liu, Ruikang and Bai, Haoli and Bao, Han and Zhao, Kang and Li, Yuening and Hu, Jiaxin and Yu, Xianzhi and Hou, Lu and Yuan, Chun and others},
  booktitle={ICML},
  year={2025}
}

@inproceedings{liu2024spinquant,
  title={Spinquant: Llm quantization with learned rotations},
  author={Liu, Zechun and Zhao, Changsheng and Fedorov, Igor and Soran, Bilge and Choudhary, Dhruv and Krishnamoorthi, Raghuraman and Chandra, Vikas and Tian, Yuandong and Blankevoort, Tijmen},
  booktitle={ICLR},
  year={2025}
}

@inproceedings{liu2025paretoq,
  title={ParetoQ: Improving scaling laws in extremely low-bit LLM quantization},
  author={Liu, Zechun and Zhao, Changsheng and Huang, Hanxian and Chen, Sijia and Zhang, Jing and Zhao, Jiawei and Roy, Scott and Jin, Lisa and Xiong, Yunyang and Shi, Yangyang and others},
  booktitle={NeurIPS},
  year={2025}
}

@inproceedings{BinaryMoS,
  title={Mixture of Scales: Memory-Efficient Token-Adaptive Binarization for Large Language Models},
  author={Dongwon Jo and Taesu Kim and Yulhwa Kim and Jae-Joon Kim},
  booktitle={NeurIPS},
  year={2024}
  }

@inproceedings{frantar_gptq_2023,
	title = {{GPTQ}: {Accurate} {Post}-{Training} {Quantization} for {Generative} {Pre}-trained {Transformers}},
	author = {Frantar, Elias and Ashkboos, Saleh and Hoefler, Torsten and Alistarh, Dan},
    booktitle ={ICLR},
	year = {2023},
}

@inproceedings{ashkboos_quarot_2024,
	title = {{QuaRot}: {Outlier}-{Free} 4-{Bit} {Inference} in {Rotated} {LLMs}},
	author = {Ashkboos, Saleh and Mohtashami, Amirkeivan and Croci, Maximilian L. and Li, Bo and Cameron, Pashmina and Jaggi, Martin and Alistarh, Dan and Hoefler, Torsten and Hensman, James},
    booktitle={NeurIPS},
	year = {2024},
}

@inproceedings{li_gptaq_2025,
	title = {{GPTAQ}: {Efficient} {Finetuning}-{Free} {Quantization} for {Asymmetric} {Calibration}},
	author = {Li, Yuhang and Yin, Ruokai and Lee, Donghyun and Xiao, Shiting and Panda, Priyadarshini},
    booktitle={ICML},
	year = {2025},
}

@inproceedings{arai_quantization_2026,
	title = {Quantization {Error} {Propagation}: {Revisiting} {Layer}-{Wise} {Post}-{Training} {Quantization}},
	author = {Arai, Yamato and Ichikawa, Yuma},
    booktitle={NeurIPS},
	year = {2025},
}

@inproceedings{huang_slim-llm_2025,
	title = {{SliM}-{LLM}: {Salience}-{Driven} {Mixed}-{Precision} {Quantization} for {Large} {Language} {Models}},
	author = {Huang, Wei and Qin, Haotong and Liu, Yangdong and Li, Yawei and Liu, Qinshuo and Liu, Xianglong and Benini, Luca and Magno, Michele and Zhang, Shiming and Qi, Xiaojuan},
    booktitle={ICML},
	year = {2025},
}

@inproceedings{kim_boa_nodate,
	title = {{BOA}: {Attention}-aware {Post}-training {Quantization} without {Backpropagation}},
	author = {Kim, Junhan and Kim, Ho-young and Cho, Eulrang and Lee, Chungman and Kim, Joonyoung and Jeon, Yongkweon},
    booktitle={ICML},
    year = {2025},
}

@inproceedings{liu_vptq_2024,
	title = {{VPTQ}: {Extreme} {Low}-bit {Vector} {Post}-{Training} {Quantization} for {Large} {Language} {Models}},
	author = {Liu, Yifei and Wen, Jicheng and Wang, Yang and Ye, Shengyu and Zhang, Li Lyna and Cao, Ting and Li, Cheng and Yang, Mao},
    booktitle={EMNLP},
	year = {2024},
}

@inproceedings{ashkboos_halo_2025,
	title = {{HALO}: {Hadamard}-{Assisted} {Lower}-{Precision} {Optimization} for {LLMs}},
	author = {Ashkboos, Saleh and Nikdan, Mahdi and Tabesh, Soroush and Castro, Roberto L. and Hoefler, Torsten and Alistarh, Dan},
    booktitle ={NeurIPS},
	year = {2025},
}

@inproceedings{lin_awq_2024,
	title = {{AWQ}: {Activation}-aware {Weight} {Quantization} for {LLM} {Compression} and {Acceleration}},
	author = {Lin, Ji and Tang, Jiaming and Tang, Haotian and Yang, Shang and Chen, Wei-Ming and Wang, Wei-Chen and Xiao, Guangxuan and Dang, Xingyu and Gan, Chuang and Han, Song},
    booktitle={MLSys},
	year = {2024},
}

@inproceedings{zellers_hellaswag_2019,
	 title={Hellaswag: Can a machine really finish your sentence?},
  author={Zellers, Rowan and Holtzman, Ari and Bisk, Yonatan and Farhadi, Ali and Choi, Yejin},
  booktitle={ACL},
  year={2019}
}

@article{clark_think_2018,
	 title={Think you have solved question answering? try arc, the ai2 reasoning challenge},
  author={Clark, Peter and Cowhey, Isaac and Etzioni, Oren and Khot, Tushar and Sabharwal, Ashish and Schoenick, Carissa and Tafjord, Oyvind},
  journal={arXiv preprint arXiv:1803.05457},
  year={2018}
}

@inproceedings{sakaguchi_winogrande_2019,
	title={WINOGRANDE: An Adversarial Winograd Schema Challenge at Scale},
  author={Sakaguchi, Keisuke and Le Bras, Ronan and Bhagavatula, Chandra and Choi, Yejin},
  booktitle={AAAI},
  year={2020}
}

@inproceedings{chakrabarty2021figurative,
  title={Figurative language in recognizing textual entailment},
  author={Chakrabarty, Tuhin and Ghosh, Debanjan and Poliak, Adam and Muresan, Smaranda},
  booktitle={ACL},
  year={2021}
}

@inproceedings{mihaylov_can_2018,
	title={Can a suit of armor conduct electricity? a new dataset for open book question answering},
  author={Mihaylov, Todor and Clark, Peter and Khot, Tushar and Sabharwal, Ashish},
  booktitle={EMNLP},
  year={2018}
}

@article{grattafiori_llama_2024,
  title={The llama 3 herd of models},
  author={Dubey, Abhimanyu and Jauhri, Abhinav and Pandey, Abhinav and Kadian, Abhishek and Al-Dahle, Ahmad and Letman, Aiesha and Mathur, Akhil and Schelten, Alan and Yang, Amy and Fan, Angela and others},
  journal={arXiv preprint arXiv:2407.21783},
  year={2024}
}

@inproceedings{merity_pointer_2016,
	title={Pointer sentinel mixture models},
  author={Merity, Stephen and Xiong, Caiming and Bradbury, James and Socher, Richard},
  booktitle={ICLR},
  year={2017}
}

@article{2020t5,
  author  = {Colin Raffel and Noam Shazeer and Adam Roberts and Katherine Lee and Sharan Narang and Michael Matena and Yanqi Zhou and Wei Li and Peter J. Liu},
  title   = {Exploring the Limits of Transfer Learning with a Unified Text-to-Text Transformer},
  journal = {JMLR},
  year    = {2020},
}

@inproceedings{hendrycks_measuring_2021,
	title={Measuring Massive Multitask Language Understanding},
  author={Dan Hendrycks and Collin Burns and Steven Basart and Andy Zou and Mantas Mazeika and Dawn Song and Jacob Steinhardt},
  booktitle={ICLR},
  year={2021}
}

@inproceedings{bisk_piqa_2019,
  title={Piqa: Reasoning about physical commonsense in natural language},
  author={Bisk, Yonatan and Zellers, Rowan and Gao, Jianfeng and Choi, Yejin and others},
  booktitle={AAAI},
  year={2020}
}

@inproceedings{du_bitdistiller_2024,
	title = {{BitDistiller}: {Unleashing} the {Potential} of {Sub}-4-{Bit} {LLMs} via {Self}-{Distillation}},
	author = {Du, Dayou and Zhang, Yijia and Cao, Shijie and Guo, Jiaqi and Cao, Ting and Chu, Xiaowen and Xu, Ningyi},
    booktitle={ACL},
	year = {2024},
}

@inproceedings{chen_efficientqat_2025,
	title = {{EfficientQAT}: {Efficient} {Quantization}-{Aware} {Training} for {Large} {Language} {Models}},
	author = {Chen, Mengzhao and Shao, Wenqi and Xu, Peng and Wang, Jiahao and Gao, Peng and Zhang, Kaipeng and Luo, Ping},
    booktitle={ACL},
	year = {2025},
}

@inproceedings{liu_llm-qat_2023,
	title = {{LLM}-{QAT}: {Data}-{Free} {Quantization} {Aware} {Training} for {Large} {Language} {Models}},
	author = {Liu, Zechun and Oguz, Barlas and Zhao, Changsheng and Chang, Ernie and Stock, Pierre and Mehdad, Yashar and Shi, Yangyang and Krishnamoorthi, Raghuraman and Chandra, Vikas},
    booktitle={ACL},
	year = {2024},
}

@inproceedings{shao_omniquant_2024,
	title = {{OmniQuant}: {Omnidirectionally} {Calibrated} {Quantization} for {Large} {Language} {Models}},
	author = {Shao, Wenqi and Chen, Mengzhao and Zhang, Zhaoyang and Xu, Peng and Zhao, Lirui and Li, Zhiqian and Zhang, Kaipeng and Gao, Peng and Qiao, Yu and Luo, Ping},
    booktitle={ICLR},
	year = {2024},
}

@inproceedings{shang_pb-llm_2023,
	title = {{PB}-{LLM}: {Partially} {Binarized} {Large} {Language} {Models}},
	author = {Shang, Yuzhang and Yuan, Zhihang and Wu, Qiang and Dong, Zhen},
    booktitle={ICLR},
	year = {2024},
}

@article{wang_bitnet_2023,
	title = {{BitNet}: {Scaling} 1-bit {Transformers} for {Large} {Language} {Models}},
	author = {Wang, Hongyu and Ma, Shuming and Dong, Li and Huang, Shaohan and Wang, Huaijie and Ma, Lingxiao and Yang, Fan and Wang, Ruiping and Wu, Yi and Wei, Furu},
    journal={arXiv preprint arXiv:2310.11453},
	year = {2023},
}

@inproceedings{Xu2024OneBitTE,
  title={OneBit: Towards Extremely Low-bit Large Language Models},
  author={Yuzhuang Xu and Xu Han and Zonghan Yang and Shuo Wang and Qingfu Zhu and Zhiyuan Liu and Weidong Liu and Wanxiang Che},
    booktitle={NeurIPS},
  year={2024},
}

@inproceedings{wei_qdrop_2023,
	title = {{QDrop}: {Randomly} {Dropping} {Quantization} for {Extremely} {Low}-bit {Post}-{Training} {Quantization}},
	author = {Wei, Xiuying and Gong, Ruihao and Li, Yuhang and Liu, Xianglong and Yu, Fengwei},
    booktitle={ICLR},
	year = {2022},
}

@inproceedings{lee_owq_2024,
	title = {{OWQ}: {Outlier}-{Aware} {Weight} {Quantization} for {Efficient} {Fine}-{Tuning} and {Inference} of {Large} {Language} {Models}},
	author = {Lee, Changhun and Jin, Jungyu and Kim, Taesu and Kim, Hyungjun and Park, Eunhyeok},
    booktitle={AAAI},
	year = {2024},
}

@inproceedings{xiao_smoothquant_2024,
	title = {{SmoothQuant}: {Accurate} and {Efficient} {Post}-{Training} {Quantization} for {Large} {Language} {Models}},
	author = {Xiao, Guangxuan and Lin, Ji and Seznec, Mickael and Wu, Hao and Demouth, Julien and Han, Song},
    booktitle={ICML},
	year = {2023},
}

@inproceedings{kim_squeezellm_2024,
	title = {{SqueezeLLM}: {Dense}-and-{Sparse} {Quantization}},
	author = {Kim, Sehoon and Hooper, Coleman and Gholami, Amir and Dong, Zhen and Li, Xiuyu and Shen, Sheng and Mahoney, Michael W. and Keutzer, Kurt},
    booktitle={ICML},
	year = {2024},
}

@inproceedings{chee_quip_2024,
	title = {{QuIP}: 2-{Bit} {Quantization} of {Large} {Language} {Models} {With} {Guarantees}},
	author = {Chee, Jerry and Cai, Yaohui and Kuleshov, Volodymyr and Sa, Christopher De},
    booktitle={NeurIPS},
	year = {2023},
}

@inproceedings{dettmers_spqr_2023,
	title = {{SpQR}: {A} {Sparse}-{Quantized} {Representation} for {Near}-{Lossless} {LLM} {Weight} {Compression}},
	author = {Dettmers, Tim and Svirschevski, Ruslan and Egiazarian, Vage and Kuznedelev, Denis and Frantar, Elias and Ashkboos, Saleh and Borzunov, Alexander and Hoefler, Torsten and Alistarh, Dan},
    booktitle={ICLR},
	year = {2024},
}

@inproceedings{yao_zeroquant_2022,
	title = {{ZeroQuant}: {Efficient} and {Affordable} {Post}-{Training} {Quantization} for {Large}-{Scale} {Transformers}},
	author = {Yao, Zhewei and Aminabadi, Reza Yazdani and Zhang, Minjia and Wu, Xiaoxia and Li, Conglong and He, Yuxiong},
    booktitle={NeurIPS},
	year = {2022},
}

@inproceedings{tseng_quip_2024,
	title = {{QuIP}\#: {Even} {Better} {LLM} {Quantization} with {Hadamard} {Incoherence} and {Lattice} {Codebooks}},
	author = {Tseng, Albert and Chee, Jerry and Sun, Qingyao and Kuleshov, Volodymyr and Sa, Christopher De},
    booktitle={ICML},
	year = {2024},
}

@inproceedings{tseng_qtip_2025,
	title = {{QTIP}: {Quantization} with {Trellises} and {Incoherence} {Processing}},
	author = {Tseng, Albert and Sun, Qingyao and Hou, David and Sa, Christopher De},
    booktitle={NeurIPS},
	year = {2024},
}

@inproceedings{baalen_gptvq_2025,
	title = {{GPTVQ}: {The} {Blessing} of {Dimensionality} for {LLM} {Quantization}},
	author = {Baalen, Mart van and Kuzmin, Andrey and Koryakovskiy, Ivan and Nagel, Markus and Couperus, Peter and Bastoul, Cedric and Mahurin, Eric and Blankevoort, Tijmen and Whatmough, Paul},
  booktitle = {Workshop on Efficient Systems for Foundation Models II @ ICML},
  year      = {2024}
}

@article{qwen3technicalreport,
  title     = {Qwen3 Technical Report},
    author = {An Yang and Anfeng Li and Baosong Yang and Beichen Zhang and Binyuan Hui and Bo Zheng and Bowen Yu and Chang Gao and Chengen Huang and Chenxu Lv},
  journal   = {arXiv preprint arXiv:2505.09388},
  year      = {2025},
}

@article{zhang_opt_2022,
	title = {{OPT}: {Open} {Pre}-trained {Transformer} {Language} {Models}},
	author = {Zhang, Susan and Roller, Stephen and Goyal, Naman and Artetxe, Mikel and Chen},
    journal= {arXiv preprint arXiv:2205.01068},
	year = {2022},
}

@article{krishnamoorthi2018quantizing,
  title={Quantizing deep convolutional networks for efficient inference: A whitepaper},
  author={Raghuraman Krishnamoorthi},
  journal={arXiv preprint arXiv:1806.08342},
  year={2018}
}

@inproceedings{esser_lsq_2019,
  title={Learned Step Size Quantization},
author = {Steven K. Esser and Jeffrey L. McKinstry and Deepika Bablani and Rathinakumar Appuswamy and Dharmendra S. Modha},
  booktitle={ICLR},
  year={2020}
}

@inproceedings{li_norm_2023,
	title = {Norm {Tweaking}: {High}-performance {Low}-bit {Quantization} of {Large} {Language} {Models}},
	author = {Li, Liang and Li, Qingyuan and Zhang, Bo and Chu, Xiangxiang},
    booktitle={AAAI},
	year = {2024},
}

@inproceedings{paszke2019pytorch,
  author    = {Adam Paszke and Sam Gross and Francisco Massa and Adam Lerer and James Bradbury and Gregory Chanan and Trevor Killeen and Zeming Lin and Natalia Gimelshein and Luca Antiga and et al.},
  title     = {PyTorch: An imperative style, high-performance deep learning library},
  booktitle = {NeurIPS},
  year      = {2019}
}

@inproceedings{wolf-etal-2020-transformers,
    title = "Transformers: State-of-the-Art Natural Language Processing",
    author = "Thomas Wolf and Lysandre Debut and Victor Sanh and Julien Chaumond and Clement Delangue and Anthony Moi and Pierric Cistac and Tim Rault and Rémi Louf and Morgan Funtowicz and Joe Davison and Sam Shleifer and Patrick von Platen and Clara Ma and Yacine Jernite and Julien Plu and Canwen Xu and Teven Le Scao and Sylvain Gugger and Mariama Drame and Quentin Lhoest and Alexander M. Rush",
    booktitle = "EMNLP",
    year = "2020",
}
\bibliographystyle{icml2026}

%%%%%%%%%%%%%%%%%%%%%%%%%%%%%%%%%%%%%%%%%%%%%%%%%%%%%%%%%%%%%%%%%%%%%%%%%%%%%%%
%%%%%%%%%%%%%%%%%%%%%%%%%%%%%%%%%%%%%%%%%%%%%%%%%%%%%%%%%%%%%%%%%%%%%%%%%%%%%%%
% APPENDIX
%%%%%%%%%%%%%%%%%%%%%%%%%%%%%%%%%%%%%%%%%%%%%%%%%%%%%%%%%%%%%%%%%%%%%%%%%%%%%%%
%%%%%%%%%%%%%%%%%%%%%%%%%%%%%%%%%%%%%%%%%%%%%%%%%%%%%%%%%%%%%%%%%%%%%%%%%%%%%%%
% \newpage
% \appendix
% \onecolumn
% \section{Appendix.}

\end{document}